\pdfoutput=1

\documentclass[11pt]{article}

\usepackage{EMNLP2023}

\usepackage{times}
\usepackage{latexsym}

\usepackage[T1]{fontenc}

\usepackage[utf8]{inputenc}

\usepackage{microtype}

\usepackage{inconsolata}

\usepackage{algorithm}
\usepackage{algorithmic}
\usepackage{multirow}
\usepackage{float}
\usepackage{booktabs}
\usepackage{graphicx}

\title{Investigating Antigram Behaviour using Distributional Semantics}

\author{Saptarshi Sengupta \\ The Pennsylvania State University \\
  \texttt{sks6765@psu.edu}}

\begin{document}
\maketitle
\begin{abstract}
The field of computational linguistics constantly presents new challenges and topics for research. Whether it be analyzing word usage changes over time or identifying relationships between pairs of seemingly unrelated words. To this point, we identify Anagrams and Antigrams as words possessing such unique properties. The presented work is an exploration into generating anagrams from a given word and determining whether there exists antigram (semantically opposite anagrams) relationships between the pairs of generated anagrams using GloVe embeddings. We propose a rudimentary, yet interpretable, rule-based algorithm for detecting antigrams. On a small dataset of just 12 antigrams, our approach yielded an accuracy of 39\% which shows that there is much work left to be done in this space.
\end{abstract}

\section{Introduction}

An anagram can be defined as a kind of word play where all the characters in the word are rearranged, using each character exactly once in order to generate a new word (s) which may or may not share semantic relationships with the root word. `Live' and `Vile' are examples of anagrams. Sometimes it is also possible to create a number of words from the root word using anagramming so as to produce a phrase; such is with the case `Dormitory' and `Dirty Room'. Anagrams are present in a multitude of domains ranging from literature (a famous example is seen in the novel \textit{The Da Vinci Code} where the phrase \textit{O, Draconian devil!} was an anagram of \textit{Leonardo Da Vinci}) to cyber security (for solving certain kinds of cryptograms such as the transposition and permutation ciphers). Antigrams on the other hand, are a class of anagrams which share an antonymic relationship with their anagram partner. For instance, `medicate' and `decimate' are examples of antigrams. These words are much more interesting to study because instead of a simple word play used to generate new phrases (anagrams) which \textit{might} have a connection to the original word, the task can be to find a new word from the root word with which it has a strong relationship (antonym). Such exploration, while simple, can be used to probe the reasoning abilities of Large Language Models for example, as a task in benchmarks such as BigBench \cite{srivastava2023beyond} which includes such games as the \textit{high\_low\_game} (think of a number and ask the language model to guess it by responding higher/lower than the guess).

In light of Natural Language Processing (NLP) tasks, very little work has been performed on antonyms and even less work has been done on anagrams and antigrams. Most of the work done has been conducted from the viewpoint of psychology experiments where researchers try to understand cognitive processing in the human mind. \citet{adams2011role} presented a study showing how the number of syllables influenced the difficulty of solving an anagram for both skilled and unskilled problem solvers. A similar task was undertaken by \citet{novick2008effects}. \citet{vincent2006anagram} created software which enabled users to discover novel anagrams and classify existing anagrams on the basis of certain psycholinguistic variables. Anagrams also help in understanding how cognition is linked with age or personality changes \cite{java1992priming}. \citet{thalenberg2016distinguishing} conducted research in exploring antonyms and distinguishing their presence in vector space by using Word2Vec embeddings \cite{DBLP:journals/corr/abs-1301-3781}. But as mentioned before, none of the above works truly examined anagram or antigram relationships among words from the standpoint of NLP application. Rather they were explorations in cognitive science.

Generating single word anagrams is a relatively trivial task, simply compute every possible permutation of the characters of the given word and eliminate those terms which are noisy i.e. not found in either a corpus or a dictionary. However, trying to automatically detect antigrams among the generated anagrams in a trickier task since semantic information is required before such analysis can be conducted. This idea forms the main motivation of our work. We wanted to explore how antigrams are related to each other in terms of semantic similarity using word embedding models viz. GloVe embeddings, or in other words, automatically determine pairs of antigrams from the anagrams of a given word. The results obtained were compared with GloVe similarity scores computed between well-known antonyms and it was found that pure antonyms have, on average, higher similarity to each other than pairs of antigrams.

The rest of the paper is organized as follows. Section \ref{sec:prop_meth} describes the methodology of our work. Section \ref{sec:RA} provides the results and analysis of experimentation. Finally, the paper is concluded and future directions are described in section \ref{sec:conc}.

\section{Proposed Methodology}
\label{sec:prop_meth}

The key question which our system aims to address is whether pairs of anagrams generated from a target word share an antigram relationship. To solve this question, we devise a simple algorithm (cf. algorithm \ref{alg:anagram_antigram}) which classifies a pair of words (anagrams) as having or not having antigram relationship on the basis of a threshold value which is set empirically.

\begin{algorithm}
\caption{Anagram generator and Antigram checker}
\label{alg:anagram_antigram}
\begin{algorithmic}[1]
\REQUIRE Root word $W$
\ENSURE List of anagrams of $W$ and those pairs which are antigrams
\STATE Generate every permutation of all the letters in $W$
\FORALL{$P$ in permutations}
    \IF{$P$ is not a valid lexical form}
        \STATE Remove $P$ from permutations
    \ENDIF
\ENDFOR
\STATE Print permutations $\triangleleft$ This is the Anagram List
\STATE Set antigram\_list $\leftarrow$ []
\FORALL{pair $C$ in permutations}
    \STATE $z \leftarrow$ sim($C_0$, $C_1$)
    \IF{$z \leq -0.06$}
        \STATE Add $C$ to antigram\_list
    \ENDIF
\ENDFOR
\RETURN antigram list
\end{algorithmic}
\end{algorithm}


Our work makes use of the GloVe model \cite{pennington2014glove} for calculating semantic similarity between the generated anagrams. GloVe, one of the earliest attempts at creating dense word representations, is an unsupervised learning model which learns real numbered vectors, by analyzing word co-occurrences in a given corpus. What differentiates GloVe from its predecessor viz. Word2Vec is the \textit{local} v/s \textit{global} context scope in learning embeddings i.e. while Word2Vec relies on generating pairs of positive and negative samples from the corpus, using a given window size (thus local context) GloVe learns its embeddings from a big word co-occurrence matrix it constructs over the entire corpus (thus global view) and optimizes for a word association probability objective function.

In the proposed algorithm, each permutation (\textit{P}) of the root word (\textit{W}) is run through a spellchecker which was implemented in our work with the help of the \textit{PyEnchant}\footnote{\url{https://pythonhosted.org/pyenchant/}} package for Python. A spell-checking module was required so as to filter out the invalid lexical forms of \textit{W} (as all the permutations would not be proper English words). The members in the filtered permutation list are the anagrams that were required. Using their respective GloVe embedding, semantic similarity between each unique pair (\textit{C}) of anagrams from the filtered permutation list was computed. The members of pair \textit{C} were termed as $C_0$ and $C_1$ . Finally, if the similarity score between $C_0$ and $C_1$ was found to be less than -0.06, the pair was selected as an antigram. We set the value to -0.06 for two reasons, 

\begin{itemize}
    \item Assuming antonym vectors point $180^{\circ}$ w.r.t each other, the cosine similarity in turn would be a negative score [$\cos(180^{\circ}) = -1$].
    \item We arrived at 0.06 by taking an average of all the scores obtained during our trials.
\end{itemize}

A point of contention regarding the algorithm would be, the exclusive use of GloVe vectors for generating similarity scores between the anagram pairs. We provide three reasons for this choice,

\begin{itemize}
    \item While contextualized embeddings from models such as BERT \cite{devlin-etal-2019-bert} and GPT-3 \cite{brown2020language} have been shown to outperform non-contextualized representations such as GloVe and Word2Vec, they typically tend to break down words to subwords during tokenization. As such, obtaining a unified representation for the whole word becomes non-trivial as there are a few choices to be considered, taking a mean of the subword embeddings, using the [CLS] token, etc.
    \item Additionally, given the very nature of contextualized embeddings i.e. learning representations of tokens on the fly given the surrounding context does not make it suitable when we want to analyze words and concepts in \textit{isolation} i.e. how related or unrelated their vectors are with respect to other terms without the influence of guiding context.
    \item Finally, the main purpose of the algorithm was to examine the nature of word vectors to see whether antonym relationship was equivalent to word vectors being opposite in direction in semantic space. As such, we hypothesized that the choice of embedding technique \textit{should not} matter that much if the core assumptions for vector behavior hold (cf. \ref{sec:RA}).
\end{itemize}

\section{Results Analysis}
\label{sec:RA}

We compute semantic similarity using pre-trained GloVe embeddings, trained on the Wikipedia and Gigaword\footnote{\url{https://catalog.ldc.upenn.edu/LDC2003T05}} corpus. We use the 100-dimension variant vectors.

For testing our algorithm, the antigram dataset compiled by \citet{anil2010antigrams} was taken which consists of 50 pairs of well-known antigrams. This was the only reliable source of antigrams we could obtain seeing as it is such a niche area of study. However, we could only use 12 out of these 50 since the rest were multi-word antigrams and we wanted to focus our attention on single words rather than phrases, for the reasons mentioned in sec. \ref{sec:prop_meth} and also because phrases as a whole would not be present in GloVe's vocabulary. Table \ref{tab:antigram_scores} presents the results of the antigram tests.


From the results of Table \ref{tab:antigram_scores}, it became clear that a similarity score lower than -0.06 cannot be the only criteria for an anagram pair to be declared an antigram. With an overall accuracy of 39\%, considering all 28 cases, a total of 3 pairs, out of 12 (25\%), were actual (true) antigrams, which is quite low in modern architecture standards. Additionally, \textbf{our model declared that a pair of anagrams were not antigrams more times than they were} (16 v/s 12). This could be attributed to the low similarity threshold we set. Furthermore, our approach could not identify the other antigrams for two reasons, i) either the word does not exist in the models' vocabulary (those pairs that are marked by N/A) or, ii) the similarity between the words is greater than -0.06. However, despite the poor performance, perhaps the only saving grace for our approach lies in its simplicity which leads to a more interpretable method allowing us to pinpoint where the decision error comes from.

Similarity scores were computed between well-known antonym pairs and a striking observation was made. Table \ref{tab:antonym_scores} shows the corresponding results. The average semantic similarity score for the antonym pairs (cf. Table \ref{tab:antonym_scores}) was 0.75. Such a high score is actually counterintuitive. This is because the hypothesis surrounding regarding word vectors is that if they are antonyms, they would point in opposite directions and as such the cosine of the angle between them would be negative. Such a notion is clearly challenged when the scores from the antigram tests are contrasted with the antonym tests. The scores from Table \ref{tab:antonym_scores} highlight the fact that it is not necessary for a word pair to have a negative similarity score, in order to be antonyms and vice versa. 
GloVe tries to reduce the angle between vectors of similar words and as such they become clustered very near to each other in vector space. Smaller the angle, closer is its cosine to 1. Thus, having a negative similarity score is not indicative of antonyms. Rather it was found that antonyms are strongly related to each other and as such produce high similarity scores. This fact was challenged by the antigrams whose similarity scores were extremely low and even negative in some cases (cf. Table \ref{tab:antigram_scores}) but still had an antonymic relationship. This indicates that antigrams and antonyms behave differently in semantic space in spite of sharing a common link. Extending the above logic, it could also mean that words that are considered antigrams are typically not found to share similar contexts thus resulting in lower similarity scores.

\begin{table*}[ht]
    \centering
    \begin{tabular}{cccccc}
    \toprule
    \textbf{Word}     & \textbf{Anagram}  & \textbf{Anagram-Pair}         & \textbf{Similarity Score} & \textbf{System Antigram} & \textbf{True Antigram} \\
    \midrule
    Abet     & Bate     & (abet, bate)           & -0.02            & No              & No            \\
             & Beat     & (abet, beat)           & -0.1             & Yes             & Yes           \\
             & Beta     & (abet, beta           & 0.13             & No              & No            \\
             &          & (bate, beat)           & 0.19             & No              & No            \\
             &          & (bate, beta)           & -0.1             & Yes             & No            \\
             &          & (beat, beta)           & 0.01             & No              & No            \\
    Fakir    & Kafir    & (fakir, kafir)         & 0.21             & No              & Yes           \\
    Indeed   & Denied   & (indeed, denied)       & 0.43             & No              & Yes           \\
    Interim  & Termini  & (interim, termini)     & -0.08            & Yes             & Yes           \\
             & Minteri  & (interim, mintier)     & 0                & N/A             & No            \\
             &          & (termini, mintier)     & 0                & N/A             & No            \\
    OK       & KO       & (ok, ko)               & 0.29             & No              & Yes           \\
    Rousing  & Souring  & (rousing, souring)     & 0.08             & No              & Yes           \\
    Ruthless & Hurtless & (ruthless, hurtless)   & 0                & N/A             & Yes           \\
             & Hustlers & (ruthless, hustlers)   & 0.22             & No              & No            \\
             &          & (hurtless, hustlers)   & 0                & N/A             & No            \\
    Sainted  & Instead  & (sainted, instead)     & -0.29            & Yes             & No            \\
             & Stained  & (sainted, stained)    & 0.20             & No              & No            \\
             & Detains  & (sainted, detains)     & -0.13            & Yes             & No            \\
             &          & (stained, detains)     & 0.08             & No              & Yes           \\
             &          & (stained, instead)     & -0.16            & Yes             & No            \\
             &          & (instead, detains)     & -0.06            & Yes             & No            \\
    Scolded  & Coddles  & (scolded, coddles)     & 0.19             & No              & Yes           \\
    Sheared  & Headers  & (sheared, headers)     & 0.24             & No              & No            \\
             & Adheres  & (sheared, adheres)     & -0.05            & Yes             & Yes           \\
             &          & (headers, adheres)     & -0.04            & No              & No            \\
    Striking & Skirting & (striking, skirting)   & 0.14             & No              & Yes           \\
    Tip      & Pit      & (tip, pit)             & 0.27             & No              & Yes           \\
    \bottomrule
    \end{tabular}
    \caption{Semantic Similarity Scores for Antigram Testing}
    \label{tab:antigram_scores}
\end{table*}

\begin{table}[H]
\centering
\caption{Semantic Similarity Scores for Antonym Pairs}
\label{tab:antonym_scores}
\begin{tabular}{cc}
\toprule
\textbf{Antonym} & \textbf{Similarity Score} \\
\midrule
Up-Down & 0.92 \\
Large-Small & 0.93 \\
Top-Bottom & 0.59 \\
Happy-Sad & 0.68 \\
Heavy-Light & 0.64 \\
\bottomrule
\end{tabular}
\end{table}

\section{Conclusion and Future Work}
\label{sec:conc}

The presented work aims to analyse antigrams and anagrams from the standpoint of NLP instead of treating them as subjects falling in the realm of logology (recreational linguistics). We propose a simple technique for detecting antigrams from pairs of anagrams using cosine semantic similarity. Our work is perhaps one of the first to explore this topic using distributional semantics. However, as can be seen from the paper, further research remains to be done. Firstly, figuring out a way to use contextual representations instead of static ones, which in theory should provide better context for computing similarity. Second, a more sophisticated approach to this problem could be to frame it as a text classification task. We could then take advantage of much larger models such as LLaMA \cite{touvron2023llama} which have demonstrated impressive zero/few-shot performance across a range of tasks. Ultimately, in this paper, we propose basic ideas in the direction of anagram and antigram research and hope that future developments are undertaken towards it with implications for probing the reasoning abilities of large language models.

\section*{Limitations}

Of course, there are many limitations with our work. Considering this as a pilot study in exploring the antigram phenomena, we identify three initial drawbacks, i) We focus only on unigram or single-word anagrams for detecting antigrams since GloVe cannot readily handle phrasal terms (needs further processing as mentioned before such as additive/mean of constituent embeddings) which limited us from using the entire dataset, ii) In this work we only relied on non-contextualized vectors. While they worked for our purposes here, it would certainly be interesting to see how we can leverage newer contextualized models such as BERT and GPT, and iii) We concede that the size of the dataset used was extremely small to obtain a clear picture of antigram behavior. Owing to the fact that there exists very limited resources for this task, the next undertaking can be to develop a larger and more diverse dataset for the same.

\section*{Ethics Statement}

Seeing as this work ultimately computes semantic similarity between pairs of words, we believe that the ethical questions attached to it is minimal, if at all any. However, we believe that since the choice of vectors is important here, and by extension the corpus on which they were trained, care needs to be taken to not use corpora capable of perpetuating social biases in turn yielding scores indicative of the true \textit{meaning} of the words without the influence of impeding biases.

\bibliography{emnlp2023.bib}
\bibliographystyle{acl_natbib}

\end{document}